\documentclass[nohyperref]{article}

\usepackage{microtype}
\usepackage{graphicx}
\usepackage{subfigure}
\usepackage{svg}
\usepackage{booktabs} % for professional tables

\usepackage{hyperref}
\usepackage{multirow}

\newcommand{\ie}{\textit{i}.\textit{e}.}
\newcommand{\eg}{\textit{e}.\textit{g}.}
\newcommand{\specialcell}[2][c]{%
  \begin{tabular}[#1]{@{}c@{}}#2\end{tabular}}

\usepackage[accepted]{icml2022}

\usepackage{amsmath}
\usepackage{amssymb}
\usepackage{mathtools}
\usepackage{amsthm}

\usepackage[capitalize,noabbrev]{cleveref}

\theoremstyle{plain}

\theoremstyle{definition}

\theoremstyle{remark}

\usepackage[textsize=tiny]{todonotes}

\icmltitlerunning{Attack on frequency}

\begin{document}

\twocolumn[
\icmltitle{Boosting 3D Adversarial Attacks with Attacking On Frequency}



\icmlsetsymbol{equal}{*}
\begin{icmlauthorlist}
\icmlauthor{Binbin Liu}{equal,thu}
\icmlauthor{Jinlai Zhang}{equal,gxu}
\icmlauthor{Lyujie Chen}{thu}
\icmlauthor{Jihong Zhu}{gxu,thu1}
% \icmlauthor{Iaesut Saoeu}{ed}
% \icmlauthor{Fiuea Rrrr}{to}
% \icmlauthor{Tateu H.~Yasehe}{ed,to,goo}
% \icmlauthor{Aaoeu Iasoh}{goo}
% \icmlauthor{Buiui Eueu}{ed}
% \icmlauthor{Aeuia Zzzz}{ed}
% \icmlauthor{Bieea C.~Yyyy}{to,goo}
% \icmlauthor{Teoau Xxxx}{ed}
% \icmlauthor{Eee Pppp}{ed}
\end{icmlauthorlist}

\icmlaffiliation{thu}{Department of Computer Science and Technology, Tsinghua University, Beijing, China}
\icmlaffiliation{thu1}{Department of Precision Instrument, Tsinghua University, Beijing, China}
\icmlaffiliation{gxu}{College of Mechanical Engineering, Guangxi University, Nanning, China}

\icmlcorrespondingauthor{Jihong Zhu}{jhzhu@mail.tsinghua.edu.cn}

\icmlkeywords{Machine Learning, ICML}

\vskip 0.3in
]



\printAffiliationsAndNotice{\icmlEqualContribution} % otherwise use the standard text.

\begin{abstract}
Deep neural networks (DNNs) have been shown to be vulnerable to adversarial attacks. Recently, 3D adversarial attacks, especially adversarial attacks on point clouds, have elicited mounting interest. However, adversarial point clouds obtained by previous methods show weak transferability and are easy to defend. To address these problems, in this paper we propose a novel point cloud attack (dubbed AOF) that pays more attention on the low-frequency component of point clouds. We combine the losses from point cloud and its low-frequency component to craft adversarial samples. Extensive experiments validate that AOF can improve the transferability significantly compared to state-of-the-art (SOTA) attacks, and is more robust to SOTA 3D defense methods. Otherwise, compared to clean point clouds, adversarial point clouds obtained by AOF contain more deformation than outlier.
\end{abstract}

\section{Introduction}
% \label{submission} 1.引入adv 2.大家是怎么做的 3. 指出缺点 4.提出方法
In recent years, with the widespread application of 3D vision technology in safety-critical scenarios such as autonomous driving, 3D adversarial samples have attracted more and more attention from researchers~\cite{knnattack,generatingadpoint}. The adversarial samples mislead the deep learning models by introducing small perturbation to the clean data. In the case of point cloud data, researchers mostly perform adversarial attacks on point cloud data by adding, deleting some points, or changing their coordinates~\cite{knnattack,geoa3,generatingadpoint,lggan,advpc,attsordefense}.

However, adversarial point clouds crafted by traditional 3D attacking methods (\eg, 3D-Adv~\cite{generatingadpoint}, kNN~\cite{knnattack}) often exhibit weak transferability due to overfitting to the source model. To improve the transferablity of adversarial point cloud, a few studies attempted to alleviate such overfitting by introducing some regularization, \eg, \citet{advpc} enhanced the transferability by introducing a data adversarial loss from point cloud after an autoencoder.
However, existing methods generate adversarial point cloud by indiscriminately perturbing point cloud without the awareness of intrinsic features of objects that a point cloud represent for, thus easily falling into model-specific local optimum.
As pointed out in~\cite{ilyas2019adversarialnotbug}, deep learning models learn extra ``noisy'' features together with intrinsic features of objects, while the ``noisy'' features are treated equally with object-related features to support the final decision, and such ``noisy'' features will be model-specific. However, all the SOTA 3D adversarial attack algorithms treat the intrinsic features and the ``noisy'' features equally during the generation of perturbation. 

\begin{figure}[h]
\vskip 0.2in
\begin{center}
\label{lfc_fig}
\centerline{\includegraphics[width=\columnwidth]{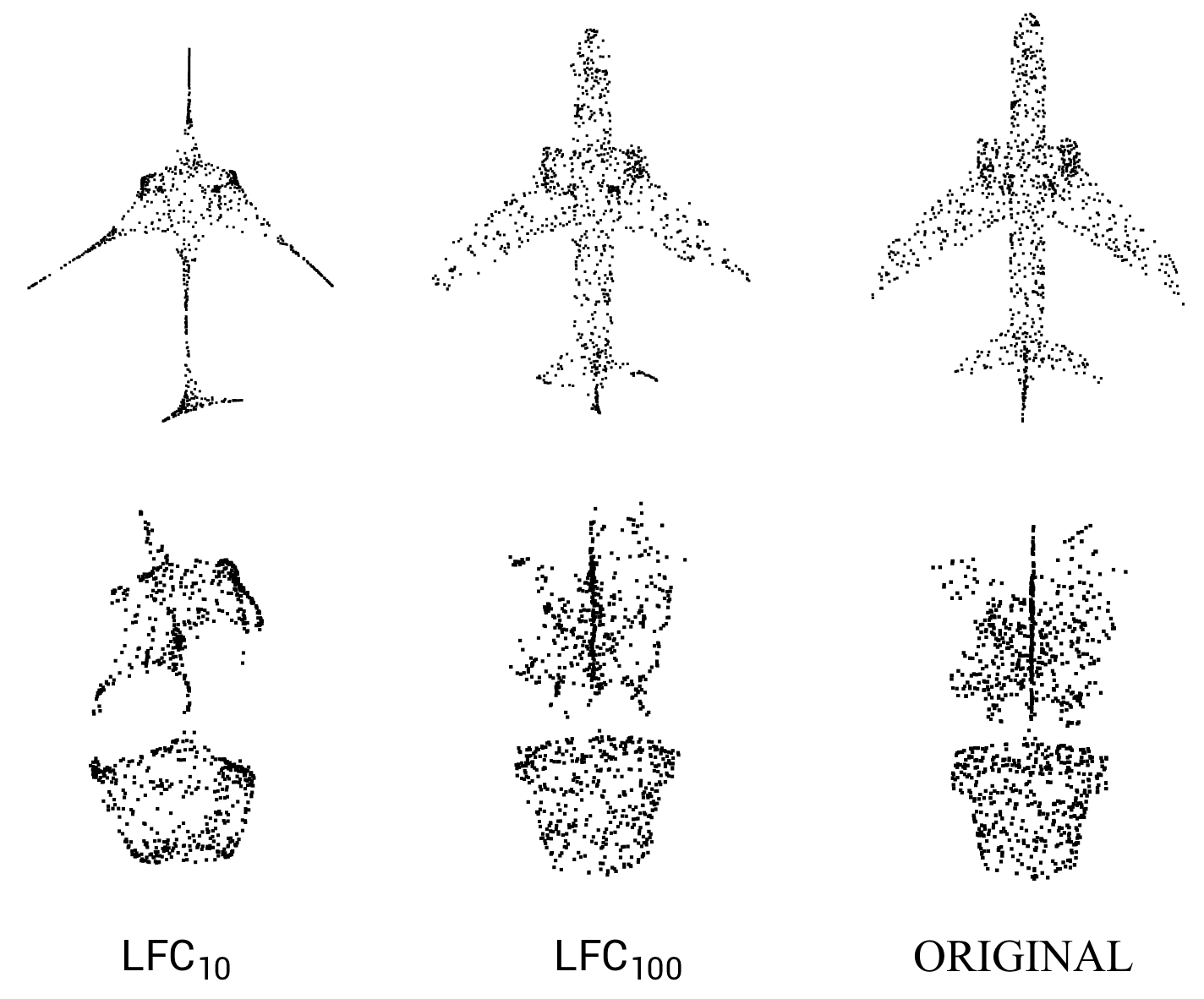}}
\caption{visualization of point cloud LFC.}
\label{pipeline}
\end{center}
\vskip -0.2in
\end{figure}

Motivated by ~\cite{wang2020high}, which explored the generalization of neural networks from the perspective of frequency in image domain. We suppose that the intrinsic features of objects are highly related to the the low-frequency component (shortened as LFC) of objects. So we make an exploration of the intrinsic features of point cloud for neural networks from the perspective of frequency. However, unlike in image domain, we cannot simply split the LFC of point cloud using discrete Fourier transform (DFT). We divide the point cloud into LFC and high-frequency component (shortened as HFC) by applying graph Fourier transform to point cloud~\cite{zhang2019frequency, xu2018cluster, shao2018hybrid, zeng20193D}. From Figure~\ref{lfc_fig}, we can see that LFC of point cloud can represents the basic shape of point cloud. So we assume that LFC plays a major role in the recognition of point cloud for the current 3D neural networks. We design a simple experiment to validate our assumption. As shown in Table ~\ref{tab:lfc-validate}, firstly we split the $\text{LFC}_{50}$, $\text{LFC}_{100}$, and $\text{LFC}_{200}$ (referred in Sec 3.1) of original point clouds, then we test the model classification accuracy on those different splitted LFCs, we found that classification accuracy only drops a little when throwing away HFC of point clouds. So we can conclude that LFC plays a major role in the recognition of point cloud, we therefore refer LFC as the intrinsic features and HFC as the ``noisy'' features.

In this paper, to improve the transferablity of adversarial point cloud, we propose Attack on Frequency (AOF) attack. We first divide a point cloud into LFC and HFC, then we pay more attention to the LFC of point cloud and add the loss of LFC to avoid overfitting to the source model. Empirical evaluation over four famous point cloud classifier demonstrates the superior performance of our AOF in terms of transferablity and robustness.

\begin{table}[t]
\caption{Classification accuracies for different point cloud classifier on various lfc components of ModelNet40 train set.}
\label{tab:lfc-validate}
\vskip 0.15in
\begin{center}
\begin{small}
%\begin{sc}
\begin{tabular}{lccccr}
\toprule
Network  & $\text{LFC}_{50}$ & $\text{LFC}_{100}$ & $\text{LFC}_{200}$ & ORIGINAL \\
\midrule
PointNet    & 92.92 & 96.54 & 97.97 & 99.84 \\
PointNet++ & 82.94 & 89.43 & 92.17 & 99.80 \\
DGCNN    & 76.34 & 88.51 & 92.02 & 100.0 \\
PointConv    & 84.15 & 93.74 & 95.72 & 99.82 \\
\bottomrule
\end{tabular}
%\end{sc}
\end{small}
\end{center}
\vskip -0.1in
\end{table}

In summary, our main contributions are given as follows:

    % 1) Unlike previous works, we add more focus on the intrinsic features of point clouds in the generation of adversarial samples through exploring 3D adversarial attack from the sight of frequency that is rarely noticed in 3D point cloud before.

    % 1): We explore the effect of low-frequency component of point cloud in the recognition of neural networks by applying graph Fourier transformation to point cloud.
    
    1) We propose a new adversarial attack AOF on point clouds which pays more attention to the LFC of point cloud and we introduce a new adversarial loss from the LFC of point cloud as a regularization in the process of optimization.
    
    2) We perform extensive experiments to validate the transferability and robustness of our attacks. These experiments show that AOF can improve the transferability significantly compared to state-of-the-art (SOTA) attacks, and is more robust to SOTA 3D defense methods.

% The guidelines below will be enforced for initial submissions and
% camera-ready copies. Here is a brief summary:
% \begin{itemize}
% \item Submissions must be in PDF\@.

% \end{itemize}

\section{Related work}

\subsection{Adversarial attacks}
The adversarial attacks are extensively investigated in recent years~\cite{fgm,ifgm,mifgm,cwattack}. They can be roughly divided into two groups, \ie, white-box attacks and black-box attacks. For white-box attack, the attackers have full knowledge of the victim model. In contrast, a black-box attack does not know any information about the victim model and can only obtain query access, \eg, the prediction output of an input. In the early works, white-box settings are popular when attacking DNNs, for example the FGSM family~\cite{fgm,ifgm,mifgm} that utilize the sign of the gradient of the victim model loss function. However, white-box attack settings are impossible to implement in practice since only query access is allowed in most realistic deep learning systems. Thus, black box attacks have received more and more attention in the adversarial machine learning community. There are two ways to perform black-box attack, one is query-based black-box attacks~\cite{al2019signblack,andriushchenko2020squareblack} which craft the adversarial samples by the response of the victim model to the given inputs. However, query-based black-box attacks suffer from  excessive queries before a successful attack, which is unacceptable from a practical standpoint. The other one is transfer-based black-box attacks~\cite{chen2020universalaoapami,mifgm}, which craft the adversarial samples via attacking a surrogate model they have white-box access to. This is promising because this approach can threaten realistic deep learning systems, \eg, Google Cloud Vision. In this paper, we focus on transfer-based black-box attack for 3D point cloud.

\subsection{3D adversarial attacks}
3D adversarial attacks aim to generate 3D adversarial samples in a human-unnoticeable way. 3D adversarial samples usually consist of two types: adversarial point cloud~\cite{generatingadpoint,knnattack} and adversarial mesh~\cite{geoa3,zhang20213Dmeshattack}. Currently, most 3D adversarial attacks are about point cloud. \cite{generatingadpoint} is the first work that crafts adversarial point cloud by adding points, small clusters or objects to clean point cloud. However, those adversarial point clouds usually contain a lot of outliers, which are not human-unnoticeable. To solve this problem, the following works~\cite{geoa3,knnattack} focus on generating adversarial point cloud with much less outliers. \citet{knnattack} proposed kNN attack that aims to generating smooth perturbation by adding chamfer distance and kNN distance to loss function as the regularization terms during optimization. \citet{geoa3} proposed the $GeoA^3$ attack which crafts adversarial point cloud in a geometric-aware way and is more human imperceptible. However, the kNN attack and the $GeoA^3$ attack are less effective in real world scenarios. Unlike previous methods, the Mesh Attack\cite{zhang20213Dmeshattack} directly adds perturbation to mesh, which can avoid the information loss during 3D printing and scanning, thus boosting the performance in real world scenarios significantly. Moreover, although the above mentioned methods achieve satisfying performance in white-box setting, it is hard for them to attack the realistic 3D deep learning systems due to their weak transferability. In order to improve the transferability of 3D adversarial attacks, AdvPC attack~\cite{advpc} adpots a point cloud autoencoder during the generation of adversarial point cloud. The experiment results show that AdvPC attack improves the transferability of 3D adversarial attacks between PointNet~\cite{pointnet}, PointNet++~\cite{pointnet++}, DGCNN~\cite{dgcnn} and PointConv~\cite{wu2019pointconv} significantly. 
In this paper, we aim to improve the transferability of 3D adversarial attacks , too.

\subsection{adversarial attacks in frequency domain}
Recently, in the image domain, there also are some works that explore the generalization and the robustness of neural networks in the frequency domain. AdvDrop ~\cite{duan2021advdrop} generates adversarial examples by dropping imperceptible components in the frequency domain. SimBA ~\cite{guo2019simple} achieves satisfactory attack success rate with an unprecedented low number of black-box queries by restricting the search to the low frequency domain. \citet{deng2020frequency} improved the performance of both white-box and black-box attacks by computing universal perturbations in the frequency domain. However, the exploration of frequency in the field of 3D point cloud adversarial attack is still lacking. This work improves the transferability and robustness of 3D point cloud adversarial attack from the view of frequency.

\section{Method}
\subsection{The frequency of point cloud}

For a point cloud $\mathcal{X}$, we construct an undirected graph $\mathcal{G}$ whose vertex set is points of $\mathcal{X}$. Each vertex connects only with its k nearest neighbors(kNN), the weight of edge between vertex $i$ and vertex $j$ is
\begin{equation}
\label{eqn:edge-weight}
    w_{ij}=
    \begin{cases}
        \exp{\left(-\frac{d^2_{ij}}{2\epsilon^2}\right)} &p_j\in \text{kNN}\left(p_j\right)\;\text{or}\;p_j\in \text{kNN}\left(p_j\right) \\
        0 &\text{otherwise,}
    \end{cases}
\end{equation}
where $\text{kNN}(p)$ represents k nearest neighbors of $p$ and $d_{ij}$ is the Euclidean distance between the two points $\mathbf{p}_i$ and $\mathbf{p}_j$,
\begin{equation}
    d_{ij} = ||\mathbf{p}_i-\mathbf{p}_j||_2
\end{equation}
With the edge weights defined above, we define the symmetric adjacency matrix $\mathbf{A}\in\mathbb{R}^{N\times N}$, with the $\left(i, j\right)$-th entry given by $w_{ij}$. $\mathbf{D}$ denotes the
diagonal degree matrix, where entry $\mathbf{D}(i,i)=\sum_{j}w_{i,j}$. The combinatorial graph Laplacian matrix is $\mathbf{L}=\mathbf{D}-\mathbf{A}$~\cite{shuman2013emerging}. $\mathbf{L}$ is symmetric and can be decomposed as
\begin{equation}
    \mathbf{L} = \mathbf{V}\mathbf{\Lambda} \mathbf{V^T}
\end{equation}
where $\mathbf{V}$ is an orthogonal matrix whose columns are eigenvectors of $\mathbf{L}$ and $\mathbf{\Lambda}$ is a diagonal matrix whose entries are the eigenvalues of $\mathbf{L}$ (the eigenvalues are sorted in an increasing order). Then, the coordinate graph signals can be projected on the eigenvectors of the Laplacian matrix $\mathbf{L}$ as follows:
\begin{equation}
\label{eqn:decomp}
    \left\{\begin{array}{ll}
    \mathbf{x} = \alpha_0\mathbf{v}_0+\alpha_1\mathbf{v}_1+\cdots+\alpha_{N-1}\mathbf{v}_{N-1}\\
    \mathbf{y} = \beta_0\mathbf{v}_0+\beta_1\mathbf{v}_1+\cdots+\beta_{N-1}\mathbf{v}_{N-1}\\
    \mathbf{z} = \gamma_0\mathbf{v}_0+\gamma_1\mathbf{v}_1+\cdots+\gamma_{N-1}\mathbf{v}_{N-1}
    \end{array}\right.
\end{equation}
where $\alpha_i=\langle\mathbf{x}{,}\mathbf{v}_i\rangle$,  $\beta_i=\langle\mathbf{y}{,}\mathbf{v}_i\rangle$ and $\gamma_i=\langle\mathbf{z}{,}\mathbf{v}_i\rangle$.

For any graph signal, we can divide it to low-frequency component (LFC) and high-frequency component (HFC) using the eigenvectors of Laplacian matrix. For example, we decompose the x-axis coordinate $\mathbf{x}=\mathbf{x}_{lfc}+\mathbf{x}_{hfc}$, with $\mathbf{x}_{lfc}$ and $\mathbf{x}_{hfc}$ defined as:
\begin{align*} 
\label{eqn:xdecomp}
    \left\{\begin{array}{l}
    \mathbf{x}_{lfc} = \alpha_0\mathbf{v}_0+\alpha_1\mathbf{v}_1+\cdots+\alpha_{m-1}\mathbf{v}_{m-1}\\
    \mathbf{x}_{hfc} = \alpha_m\mathbf{v}_m+\alpha_{m+1}\mathbf{v}_{m+1}+\cdots+\alpha_{N-1}\mathbf{v}_{N-1}
    \end{array}\right.
\end{align*}
where $m$ is a hyper-parameter that represents the number of eigenvectors in LFC. We denote the LFC with the first m eigenvectors of Laplacian matrix as $\text{LFC}_m$.

For a point cloud $\mathcal{X}$, we can divide it to low-frequency component (LFC) and high-frequency component (HFC):
\begin{equation}
\mathcal{X} = \mathcal{X}_{lfc} + \mathcal{X}_{hfc},
\end{equation}
where ${X}_{lfc}$ denotes LFC of $\mathcal{X}$ and ${X}_{hfc}$ denotes HFC of $\mathcal{X}$.

\subsection{Attack on frequency}
\begin{figure}[h]
\vskip 0.2in
\begin{center}
\centerline{\includegraphics[width=\columnwidth]{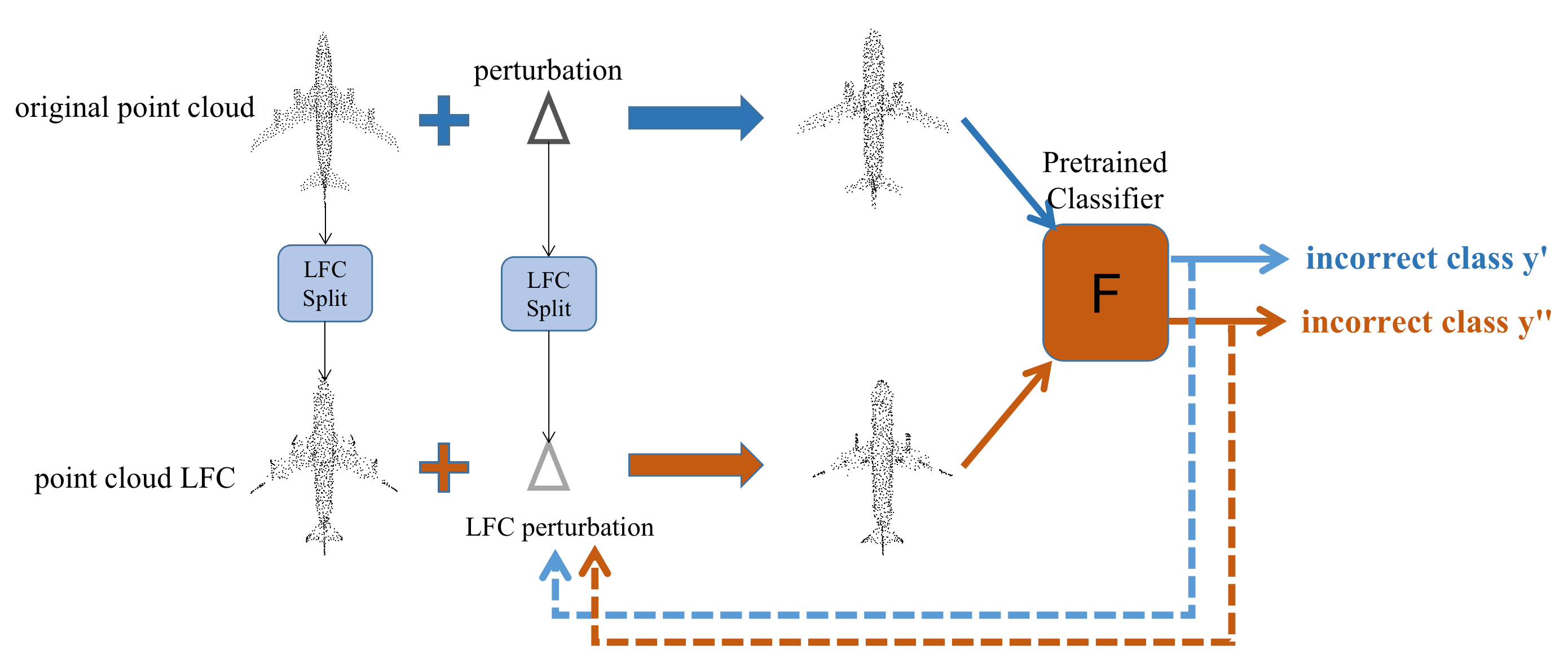}}
\caption{\textbf{AOF Attack Pipeline}: Firstly, we split the LFC of original point cloud and perturbation $\mathbf{\Delta}$. Then we get the adversarial sample $\mathcal{X}'$ and $\mathcal{X}'_{lfc}$, which is composed of the LFC of original point cloud $\mathcal{X}_{lfc}$ and the LFC perturbation $\mathbf{\Delta}_{lfc}$. The perturbed sample $\mathcal{X}'$ fools a trained classifier $\mathbf{F}$ (\ie $\mathbf{F}(\mathcal{X}')$ is incorrect), meanwhile, $\mathcal{X}'_{lfc}$ also fools the classifier $\mathbf{F}$ (\ie $\mathbf{F}(\mathcal{X}'_{lfc})$ is incorrect). We combine the original adversarial loss($blue$) and LFC adversarial loss($yellow$) to optimize the LFC perturbation $\mathbf{\Delta}_{lfc}$. Dotted lines are gradients flowing to the LFC perturbation.}
\label{pipeline}
\end{center}
\vskip -0.2in
\end{figure}

% Let the classification model $f(\cdot)$ be realized by a function $g:\mathcal{X}\rightarrow \mathbb{R}^{\mathcal{Y}}$, and
% we have $f(\mathbf{x})=\argmax_{i\in \mathcal{Y}} g_i(\mathbf{x})$, where $g_i(\cdot)$ takes the $i$th element of $\mathbf{g(\cdot)}$. When implementing $f(\cdot)$ as a deep
% classification network, $\mathbf{g(\cdot)}$ outputs the network logits, $\ie$,
% output of the network before the final softmax.

The goal of 3D adversarial attack on point cloud is to generate the adversarial point cloud example that misleads 3D point cloud classification model by adding subtle perturbation to the clean point cloud. Formally, for a classification model $\mathcal{F}:\mathcal{X} \rightarrow \mathcal{Y}$, which maps a point cloud to its corresponding class label, the adversarial examples can be generated by adding perturbation limited in a $l_{p}$ ball of size $\epsilon_p$, where $p$ can be $1$, $2$ and $\infty$. The adversarial point cloud can be represented as $\mathcal{X}' = \mathcal{X} + \mathbf{\Delta}$, where $\mathbf{\Delta}$ is the additive perturbation.

In this paper, we propose a novel 3D adversarial attack based on attacking on frequency, dubbed AOF attack. The pipeline of AOF attack is shown in Figure~\ref{pipeline}. Given a point cloud represented as $\mathcal{X} \in \mathbb{R}^{N \times 3}$, where each row is the 3D coordinates of a point. 
% Firstly, the LFC of $\mathcal{X}$ ($\mathcal{X}_{lfc}$) was obtained by using Algorithm \ref{alg:lfcsplit}. 
Firstly, we split the LFC of $\mathcal{X}$ ($\mathcal{X}_{lfc}$) using Algorithm \ref{alg:lfcsplit}. 
% Then, we get $\mathbf{\Delta}_{lfc}$ by projecting the perturbation $\mathbf{\Delta}$ on the eigenvectors
% Then, we adding perturbations in an aggressive way, we only adding perturbations on the $\mathcal{X}_{lfc}$.
Then, we get the $\mathbf{\Delta}_{lfc}$ by projecting the perturbation $\mathbf{\Delta}$ on the eigenvectors of original Laplacian matrix (Laplacian matrix of original point cloud) as follows:
\begin{equation}
    \mathbf{\Delta}_{lfc} = \left(\mathbf{v}_0, \mathbf{v}_1, \ldots, \mathbf{v}_m\right)\left(\begin{array}{cc}
         \mathbf{v}_0^T  \\
         \mathbf{v}_1^T \\
         \ldots \\
         \mathbf{v}_m^T
    \end{array}\right)\mathbf{\Delta}
\end{equation}
We represent the corresponding adversarial example of $\mathcal{X}_{lfc}$ by $\mathcal{X}_{lfc}' = \mathcal{X}_{lfc} + \mathbf{\Delta}_{lfc}$. Then we optimize the perturbation $\mathbf{\Delta}\in\mathbb{R}^{N \times 3}$ via solving the following problem:
\begin{equation}
    \min_{\mathbf{\Delta}} l_{aof}(\mathcal{X}, \mathcal{X}'), \quad s.t.D(\mathcal{X}, \mathcal{X'})\le\epsilon
\end{equation}
 As shown in Figure~\ref{pipeline}, unlike current 3D adversarial attacks, our AOF attack algorithm pays more attention to perturbing the LFC of point cloud for misleading the victim network. Moreover, we add an additional LFC adversarial loss in our AOF loss function:
 \begin{equation}
 \label{eqn:aof loss}
     l_{aof} = \left(1-\gamma\right)l_{mis}(\mathcal{X}')+\gamma l_{mis}(\mathcal{X}_{lfc}'))
\end{equation}
 
% \begin{equation}
%     \min_{\Delta} \left(1-\lambda\right)f_{t'}(F(\mathcal{X}'))+\lambda f_{t''}(F(\mathcal{X}'')),
%     s.t.D(\mathcal{X}, \mathcal{X'})\le\epsilon
% \end{equation}
where $\gamma$ is used to balance the adversarial loss between the adversarial point cloud $\mathcal{X}'$ and $\mathcal{X}_{lfc}'$. And the $l_{mis}(.)$ is defined as:
\begin{equation}
\label{eqn:margin loss}
    l_{mis}(\mathcal{X}') = \max(\max_{y \neq y_{gt}}(\mathbf{Z}(\mathcal{X}')_y) - \mathbf{Z}(\mathcal{X}')_{y_{gt}}, \kappa)
\end{equation}

% \begin{equation}
%     f_{t''}(F(\mathcal{X}'')) = \max(\max_{i \neq t''}(F(\mathcal{X}'')_i) - F(\mathcal{X}'')_{t''} + \kappa, 0)
% \end{equation}
where $y_{gt}$ is the ground truth class, $y$ is any class that not equal to $y_{gt}$, $\mathbf{Z(\cdot)}$ is the output of logits layer and $\kappa$ is the loss margin. In order to fairly compare with the kNN attack~\cite{knnattack} and the AdvPC attack~\cite{advpc}, we also follow the C\&W attack's~\cite{cwattack} optimization framework to optimize the perturbation. The AOF algorithm is summarized in Algorithm \ref{alg:aof}.
% We turn the generating adversarial point cloud to a optimization problem as follows:

% % Our adversarial loss is composed of two parts: 
% \begin{equation}
% \min _{x^{adv}} L_{M i s}\left(x^{adv}\right)+c \cdot L_{R e g}\left(x^{adv}, x\right)
% \end{equation}
% where $L_{M i s}\left(x^{adv}\right)$ can promote the misclassification of $x^{adv}$, $L_{R e g}\left(x^{adv}, x\right)$ is a regularization term that minimize the distance between $x^{adv}$ and $x$, $c$ is automated adjusted by binary search. 
% where $c$ is a trade-off between perceiveness and adversarial strength.

\begin{algorithm}[tb]
   \caption{LFCSplit}
   \label{alg:lfcsplit}
\begin{algorithmic}[1]
   \STATE {\bfseries Input:} point cloud $\mathbf{x}$, number of points $N$, number of eigenvectors in LFC $m$, number of nearest neighbors $k$
   \STATE {\bfseries Output:} low-frequency component $\mathbf{x}_{lfc}$, basis vectors of point cloud graph laplacian matrix $\mathbf{V}$;
   \STATE Connect each point with k nearest neighbors and compute the weight of each edge as formula \ref{eqn:edge-weight} to give $\mathbf{A}$;
   \STATE Initialize $\mathbf{D} = \mathbf{0}$
   \FOR{$i=0$ {\bfseries to} $N-1$}
        \STATE $\mathbf{D}\left(i, i\right) = \sum_{j}\mathbf{A}\left(i, j\right)$
   \ENDFOR
   \STATE $\mathbf{L} = \mathbf{D} - \mathbf{A}$
   \STATE Compute the eigenvectors of matrix $\mathbf{L}$ to give $\mathbf{V}$
   \STATE Initialize $\mathbf{V}_{lfc} = \mathbf{V}(:,0\cdots m-1)$ \COMMENT{//first m columns}
   \STATE $\mathbf{x}_{lfc} = \mathbf{V}_{lfc}\mathbf{V}^T_{lfc}\mathbf{x}$
\end{algorithmic}
\end{algorithm}

\begin{algorithm}[tb]
   \caption{Attack on frequency}
   \label{alg:aof}
\begin{algorithmic}[1]
   \STATE {\bfseries Input:} point cloud $\mathbf{x}$, number of points $N$, number of eigenvectors in LFC $m$, number of nearest neighbors $k$, binary search step $s$, number of iteration $n_{iter}$, balance coefficient $\gamma$, ground truth label $y_{gt}$, perturbation limit bound $\epsilon_\infty$
   \STATE {\bfseries Output:} optimized perturbation $\mathbf{\Delta}$ of point cloud $\mathbf{x}$
   \STATE Initialize $\mathbf{x}_{lfc}, \mathbf{V} = \text{LFCSplit}(\mathbf{x}, N, m, k)$
   \STATE Initialize $\mathbf{\Delta} = \mathbf{0}$
   \STATE Initialize $\mathbf{x}^0 = \mathbf{x}$
   \FOR{$step=0$ {\bfseries to} $s-1$}
        \FOR{$i=0$ {\bfseries to} $n_{iter}-1$}
            \STATE $\mathbf{V}_{lfc} = \mathbf{V}(:, 0\cdots m-1)$ \COMMENT{//first m columns}
            \STATE $\mathbf{V}_{hfc} = \mathbf{V}(:, m\cdots N-1)$ \COMMENT{//last N-m columns}
            \STATE $\mathbf{\Delta}_{lfc} = \mathbf{V}_{lfc}\mathbf{V}^T_{lfc}\mathbf{\Delta}$
            \STATE $\mathbf{\Delta}_{hfc} = \mathbf{V}_{hfc}\mathbf{V}^T_{hfc}\mathbf{\Delta}$
            \STATE $\mathbf{x}^{adv} = \mathbf{x}^0 + \mathbf{\Delta}$
            \STATE $\mathbf{x}_{lfc}^{adv} = \mathbf{x}_{lfc}^0 + \mathbf{\Delta} x_{lfc}$
            \STATE $y = F(\mathbf{x}^{adv})$
            \STATE $y_{lfc} = F(\mathbf{x}_{lfc}^{adv})$
            \STATE $loss = \left(1-\gamma\right)l_{adv}(y_{gt}, y)+\gamma l_{adv}(y_{gt}, y_{lfc})$
            \STATE $\mathbf{\Delta}_{lfc} = \min_{\mathbf{\Delta}_{lfc}}loss$
            \STATE $\mathbf{\Delta} = \mathbf{\Delta}_{lfc} + \mathbf{\Delta}_{hfc}$
            \STATE Clip $\mathbf{\Delta}$ to $\epsilon_\infty$
        \ENDFOR
   \ENDFOR
\end{algorithmic}
\end{algorithm}

\section{Experiment}
In this section, we evaluate the performance of our AOF from two aspects: the transferability compared to other SOTA attacks and the performance against SOTA point cloud defenses. The basic setups of our experiments are described in Sec 4.1. To test the efficacy of our proposed AOF, we conduct ablation studies and sensitivity analysis in Sec 4.2 and 4.3 respectively. Then we present the black-box transferability of AOF compared to other SOTA methods in Sec 4.4, and test the AOF performance under several point cloud defenses in Sec 4.5. Finally, we discuss some characteristics of AOF in Sec 4.6.
\subsection{Setup}
% In this subsection, we describe the basic setup of our experiments. 
\textbf{Dataset.} We use ModelNet40~\cite{wu20153D} to train the classifiers and test our attacks. ModelNet40 contains 12,311 CAD models from 40 different classes. These models are divided into 9,843 for training and 2,468 for testing. We use the whole test dataset to test our attacks. We trained four victim networks: PointNet~\cite{pointnet}, PointNet++~\cite{pointnet++} in Single-Scale(SSG) setting, DGCNN~\cite{dgcnn} and PointConv~\cite{wu2019pointconv}. 

\textbf{Evaluation metrics.} In the works of transfer-based black-box attack in image domain~\cite{chen2020universalaoapami,mifgm}, due to clean accuracy on the test data can reach 100\%, the transferability is evaluate by the top-1 error rate.
However, for 3D point clouds, the clean accuracy of all victim models on test dataset are around 90\%, which means there are about 10\% data we cannot ensure whether the misclassification is brouht by adversarial attacks or not. If we adopt the top-1 error rate to evaluate transferability of adversarial attack algorithms, the difference between different attacks could not be significant. We argue that the top-1 error rate cannot reflect the real performance gap between attack algorithms. Therefore, unlike previous works~\cite{advpc}, we use the percentage of the misclassified adversarial samples out of all the clean samples that are classified correctly to evaluate the transferability.
% However, for point cloud, as we point out in Table~\ref{analysistransfer} and Table~\ref{tbl:transfer}, if we use the top-1 error rate to evaluate transferability of different attack algorithms, the difference is not significant due to the clean accuracy of all victim models are around 90\%. Which could be unfair to compare the transferability of each algorithms, for example, when compare the transferability of 3D-Adv~\cite{generatingadpoint} and our AOF transfered to DGCNN~\cite{dgcnn} using top-1 error rate, the difference between 3D-Adv and AOF are 3.65 times, however, there are about 10\% data we cannot ensure the erorr is brouht by adversarial attacks. We argue that the top-1 error rate cannot reflect the real performance gap between attack algorithms, therefore, unlike previous works~\cite{advpc}, we use the percentage of the misclassified adversarial samples out of all the clean samples that are classified correctly to evaluate the transferability.

We denote the clean samples that are classified correctly by the victim model as set $\mathbf{S}$ and the corresponding adversarial samples of $\mathbf{S}$ as set $\mathbf{S}_{adv}$. In the set $\mathbf{S}_{adv}$, we denote the samples that are misclassified as set $\mathbf{T} \subseteq \mathbf{S}_{adv}$. Then the attack success rate(ASR) can be calculated by:
\begin{equation}
\label{eqn:asr}
    \text{ASR} = \frac{\left|\mathbf{T}\right|}{\left|\mathbf{S}_{adv}\right|}
\end{equation}

% We denote the clean samples that are classified correctly by the victim model as set $\mathbf{S}^{vm}$ and the corresponding adversarial samples of $\mathbf{S}^{vm}$ as set $\mathbf{S_{adv}^{vm}}$. In the set $\mathbf{S_{adv}^{vm}}$, we denote the samples that are misclassified as set $\mathbf{T}^{vm} \subseteq \mathbf{S_{adv}^{vm}}$. Then the attack success rate(ASR) can be calculated by:
% \begin{equation}
%     ASR = \frac{\left|\mathbf{T^{vm}}\right|}{\left|\mathbf{S_{adv}^{vm}}\right|}
% \end{equation}
% In terms of transferability, we define the clean samples that are classified correctly by the transfer model as set $\mathbf{S}^{tm}$, the corresponding adversarial samples craft by the victim model that are misclassified by the transfer model as set $\mathbf{S_{v2t}^{tm}}$, the transferability can be defined as follows:
% \begin{equation}
%     T_{rans} = \frac{\left|\mathbf{S_{v2t}^{tm}}\right|}{\left|\mathbf{S}^{tm}\right|} * 100\%
% \end{equation}
% To simplify the transferability to other netwoks, we also report the averaged transferability, which denotes the averaged percentage of ASR that from a victim model tranfered to all other models.

\textbf{Baselines.}
We compare AOF with the state-of-the-art baselines: 3D-Adv~\cite{generatingadpoint}, kNN attack~\cite{knnattack} and AdvPC~\cite{advpc}. For AdvPC and all of our attacks, we use Adam optimizer~\cite{kingma2014adam} with learning rate $\eta$ = 0.01, and perform 2 different initializations for the optimization of $\mathbf{\Delta}$ (as done in ~\cite{generatingadpoint}), the number of iterations for the attack optimization for all the networks is 200. We set the loss margin $\kappa$ = 30 in Eq (\ref{eqn:margin loss}) for 3D-Adv ~\cite{generatingadpoint}, AdvPC~\cite{advpc} and AOF and $\kappa$ = 15 for kNN Attack ~\cite{knnattack}(as suggested in their paper). For other hyperparameters of \cite{knnattack, generatingadpoint,advpc}, we follow the experimental details in their paper. We pick $\gamma$ = 0.25 in Eq (\ref{eqn:aof loss}) for AOF.

\subsection{Ablation Study}
In this section, we conduct a series of experiments on ModelNet40 test set to test the efficacy of AOF.

\subsubsection{hyperparameter $m$}
\begin{figure}[ht!]
\vskip 0.2in
\begin{center}
\centerline{\includegraphics[width=\columnwidth]{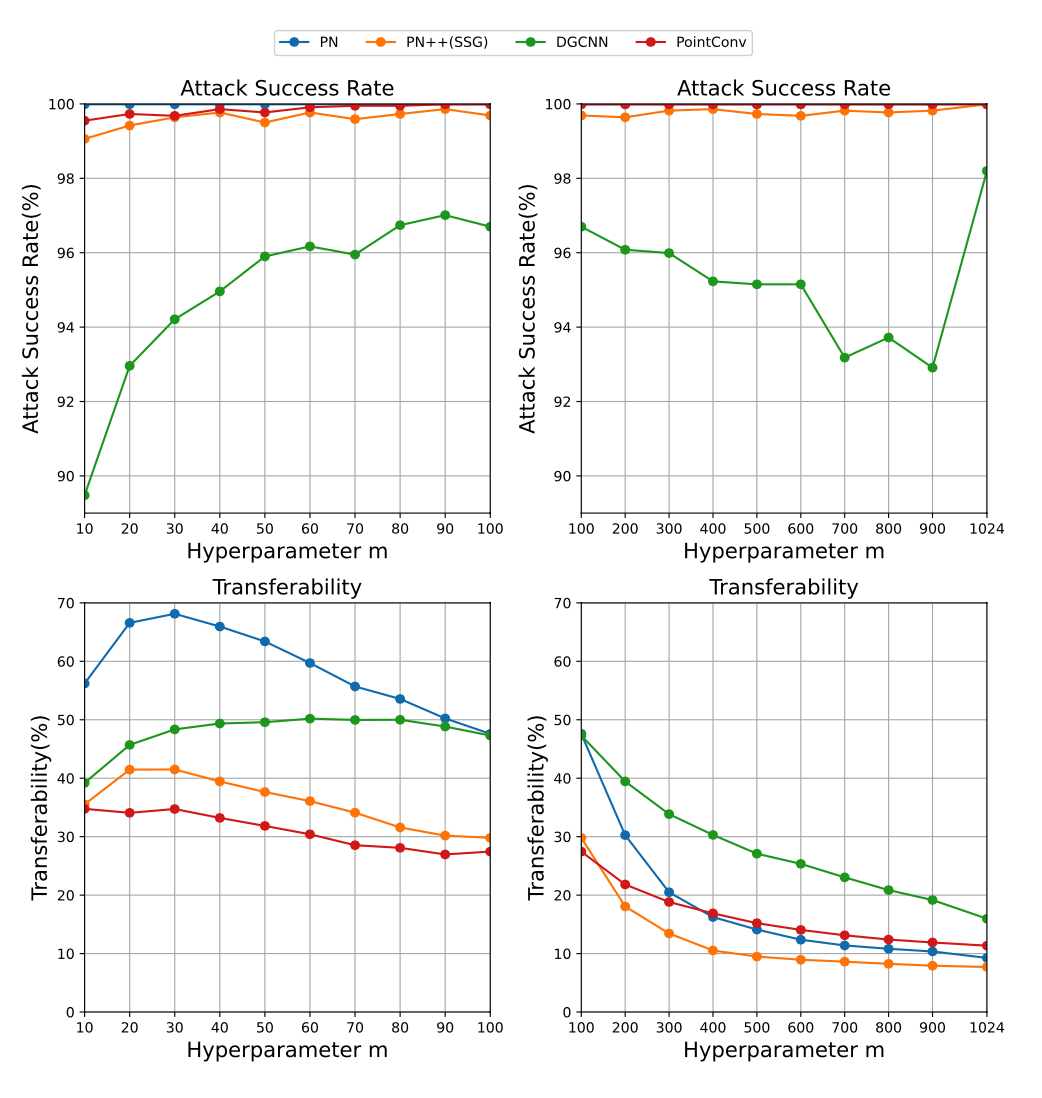}}
\caption{
% \text{Ablation study}: Studying t
The effect of changing hyperparameter $m$ on attack success rate(above) and transferability(below) of AOF. For a better view, we divide the value range of $m$ into two part:$\left[0, 100\right]$ (left) and $\left[100, 1024\right]$(right). Here, we pick $\epsilon_\infty=0.18$ and $\gamma=0.25$ and the transferability score reported for each victim network is the average attack success rate on the transfer networks.}
\label{ablation_m}
\end{center}
\vskip -0.2in
\end{figure}

Here, we study the effect of hyperparameter $m$(number of eigenvectors of Laplacian matrix in LFC) on the performance of AOF attacks. When varying $m$ between 0 and 1024, the variation of attack success rate and the averaged transferability of attacks are shown in Figure~\ref{ablation_m}, the averaged transferability denotes the averaged ASR on the transfer networks. The exact values we used for parameter $m$ are $\{10, 20, 30, 40, 50, 60, 70, 80, 90, 100, 200,$ $300, 400, 500,
600, 700, 800, 900, 1024\}$. We make the following observations.
% From Figure~\ref{ablation_m}, we observed that
(1) Except for DGCNN~\cite{dgcnn}, $m$ has little effect on the attack success rate of the victim networks.
(2) As $m$ increases from 0 to 1024, the transferability first increases and then decreases to the minimum when $m=1024$ (all frequency components). This phenomenon validates that paying more attention to LFC of point clouds can improve the transferability of attacks. But when $m$ is small, the transferability decreases for the information contained in LFC is not enough to transfer to other neural networks. We select $m=100$ to balance the transferability and attack success rate in the following experiments.

\subsubsection{hyperparameter $\gamma$}
The hyperparameter $\gamma$ is used to balance the adversarial loss between the adversarial point cloud and its LFC. The $\gamma$ is selected from $\{0.0, 0.25, 0.50, 0.75, 1.0\}$, other experiment setups are the same as Sec 4.1.

\begin{figure}[ht!]
\vskip 0.2in
\begin{center}
\centerline{\includegraphics[width=\columnwidth]{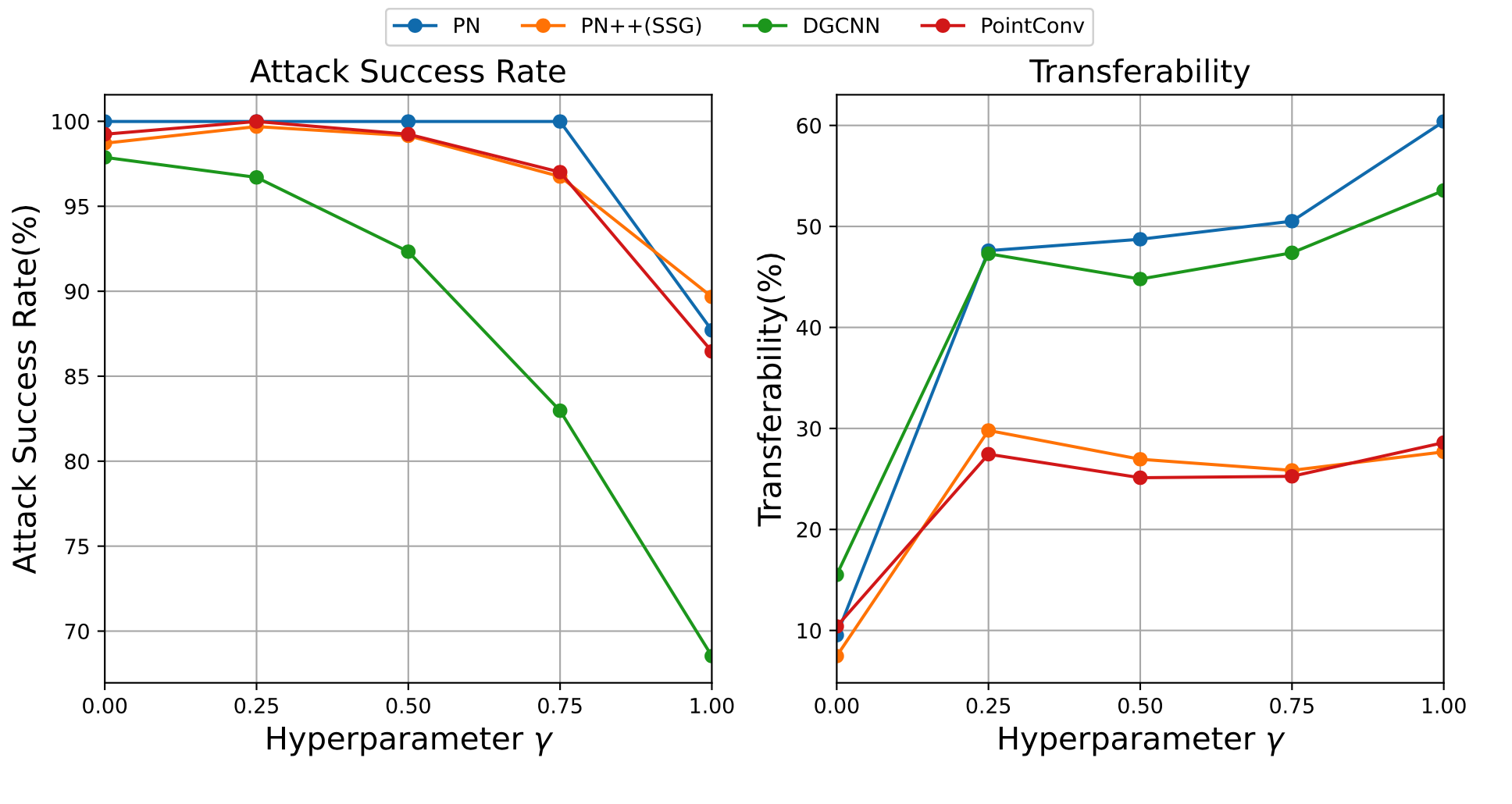}}
\caption{
% \text{Ablation study}:Studying t
The effect of changing hyperparameter $\gamma$ on attack success rate(left) and transferability(right) of AOF. Here, we pick $\epsilon_\infty=0.18$ and $m=100$. The transferability score reported for each victim network is the average of attack success rate on the transfer networks.}
\label{fig:ablation_gamma}
\end{center}
\vskip -0.2in
\end{figure}

From Figure~\ref{fig:ablation_gamma}, one observation is that adding the LFC adversarial loss with $\gamma>0$ tends to reduce the white-box attack success rate, even though it improves the transferability. Especially for DGCNN, when $\gamma$ is bigger than 0.5, the ASR decreased significantly.
%and the transferability varies with $\gamma$.
%droped firstly and inceased gradually when $\gamma$ reaching to 1.% 
We therefore pick $\gamma=0.25$ in our experiments to balance attack success rate and transferability.

\subsection{Sensitivity Analysis on Norm budget $\epsilon_\infty$}
Adversarial effects of attack methods are related with norm budget. In this section, we study the variation of attack success rate and transferability of AOF as norm budget $\epsilon_\infty$ varies. The exact values we used for parameter $\epsilon_\infty$ are $\{0.01, 0.04, 0.05, 0.08, 0.10, 0.15, 0.18\}$. Other settings are the same with Sec 4.1. 

\begin{figure}[ht!]
\vskip 0.2in
\begin{center}
\centerline{\includegraphics[width=\columnwidth]{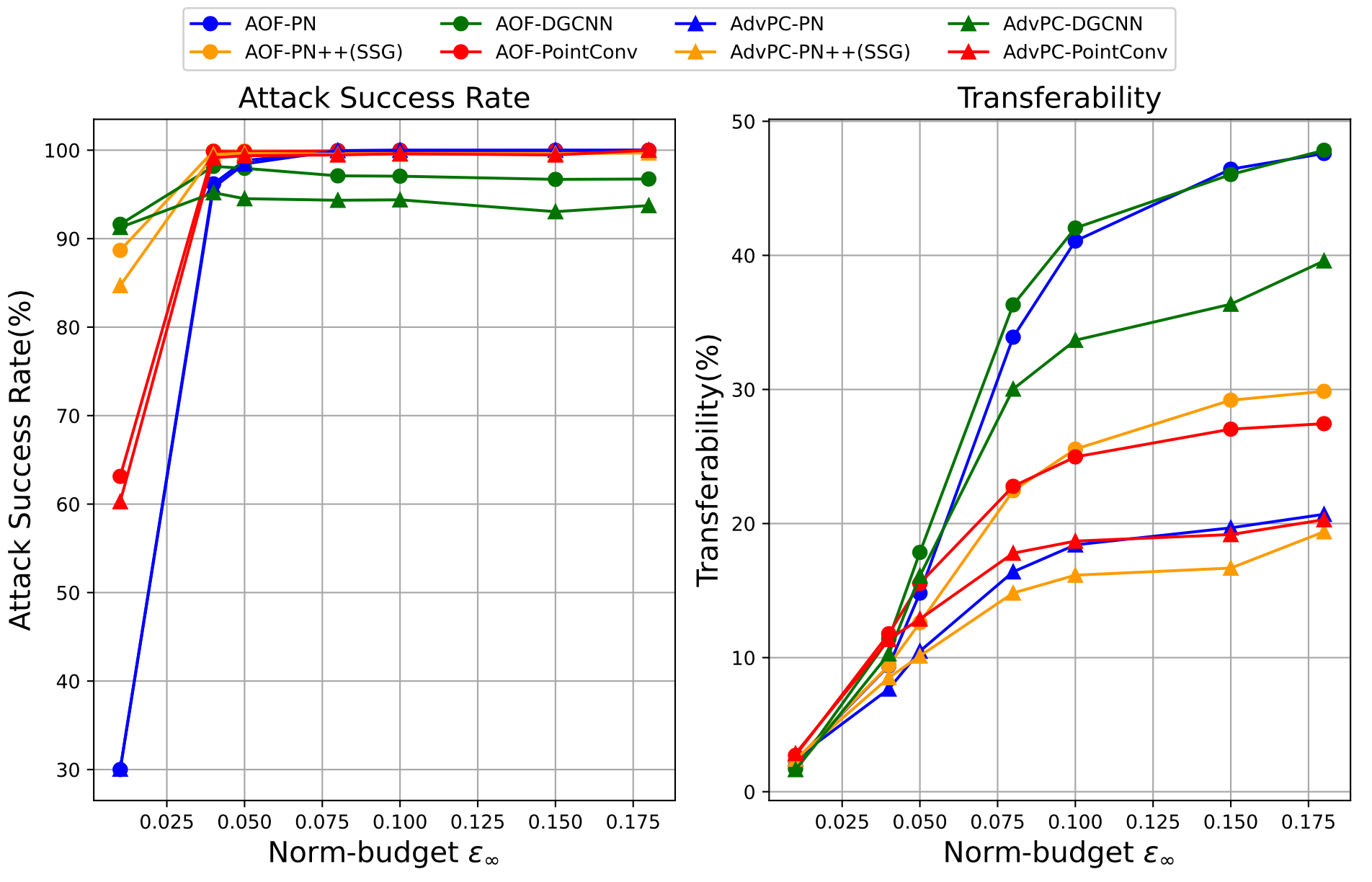}}
\caption{
% \text{Sensitivity Analysis}: Studying t
The influence of changing norm budget $\epsilon_\infty$ on attack success rate (left) and transferability (right) of AOF. Here, we pick $\gamma=0.25$ and $m=100$. The transferability score reported for each victim network is the average attack success rate on the transfer networks.}
\label{fig:budget}
\end{center}
\vskip -0.2in
\end{figure}

As shown in Figure \ref{fig:budget}, compared to transfersability, the attack success rate of AOF is less sensitive to $\epsilon_\infty$, even when $\epsilon_\infty$ is only 0.04, the attack success rate of AOF is still satisfactory. When $\epsilon_\infty$ varies in $\left[0.0, 0.18\right]$, transferability of AOF still outperforms AdvPC. As $\epsilon_\infty$ increase from 0.01 to 0.18, the transferability grows rapidly first and then grows slowly.

% The transferability grows rapidly first and then grows slowly as $\epsilon_\infty$ grows.

% The transferability of AOF outperforms AdvPC in large margin under various norm budget.

% As shown in \ref{fig:budget}, compared to the transferability, the ASR of AOF is less affected by norm budget $\epsilon_\infty$. For the transferability, the higher $\epsilon_\infty$, the transferability. Indicating that it is difficult to transfer the adversarial point cloud to other models if the attack norm budget $\epsilon_\infty$ is smaller than 0.05.

% \textbf{Targetted attack.}
% We perform targeted attack with the target label kindly provided by~\cite{wu2020ifdefense}. The results are shown in Table~\ref{target_attack}.
% We report the targeted attack success rate and classification accuracy. The results indicating the targeted attack on 3D point cloud is a challenging task. Moreover, we will explore the transferability of targeted attack in Sec 4.4.
% \begin{table}[t]
% \caption{The targetted attack on DGCNN with different norm budget $\epsilon_\infty$.}
% \label{target_attack}
% \vskip 0.05in
% \begin{center}
% \begin{small}
% \begin{sc}
% \setlength{\tabcolsep}{0.6mm}{
% \begin{tabular}{lccccccr}
% \toprule
% Index  & 0.01 & 0.04 & 0.05 & 0.08& 0.1 & 0.15 & 0.18\\
% \midrule
% % DGCNN(TR) & 0 & 0 & 0 & 0 & 0 & 0 & 0\\
% ASR & 5.35 & 27.27 & 33.43 & 38.25 & 39.79 & 39.99 & 40.80\\
% ACC & 5.63 & 18.92 & 19.77 & 15.40 & 14.30 & 9.44 & 8.83\\
% \bottomrule
% \end{tabular}}
% \end{sc}
% \end{small}
% \end{center}
% \vskip -0.1in
% \end{table}

\subsection{Transferability of AOF}
In this section, we compare our AOF with SOTA 3D adversarial attack algorithms in terms of transferability for untargetted attack following the experiment setup of ~\cite{generatingadpoint,advpc}.
%both untargetted attack and targetted attack. 

\begin{table*}[ht]
\caption{\textbf{Transferability of Attacks.} We use ASR defined in formula (\ref{eqn:asr}) to evaluate the transferability. Compared to Table~\ref{analysistransfer}, the difference of transferability to PointNet between 3D-Adv and our AOF is more obvious, which reflects the real gap between attack algorithms.
}
\vskip 0.1in
\footnotesize
\setlength{\tabcolsep}{1.7mm} %
\renewcommand{\arraystretch}{1.1} %
\centering
% \resizebox{0.9\textwidth}{!}{%
% \caption{\small \textbf{Transferability of Attacks.} Unlike Table~\ref{analysistransfer}, we use the percentage of the misclassified adversarial samples out of all the clean samples that are classified correctly to evaluate the transferability. The difference of transferability to PointNet between 3D-Adv and our AOF using ASR defined in formula \ref{eqn:asr} is more obvious, which reflects the real gap between attack algorithms. 
% }
\begin{tabular}{cc|cccc|cccc} 
\toprule
 &  &\multicolumn{4}{c|}{$\epsilon_\infty = 0.18$} &\multicolumn{4}{c}{$\epsilon_\infty = 0.45$}\\
\specialcell{Victim Networks} & Attacks  & PointNet & PointNet++ & PointConv & DGCNN  &PointNet & PointNet++  & PointConv & DGCNN   \\
\midrule
 \multirow{4}*{PointNet} & 3D-Adv  &  100 &  5.01 & 2.06 &  4.22 & 100 &  5.12 &  2.01 &  4.18 \\
 & KNN  & 100 &  11.1 & 3.62 &  10.7 & 100 &  10.8 &  3.53 &  10.2 \\
 & AdvPC  &  100 & 30.4 & 13.6 & 14.8 &  100 & 30.4 & 13.1 & 14.9 \\
  & AOF(ours)  &  99.9 & \textbf{57.2} & \textbf{36.3} & \textbf{29.4} &  100 & \textbf{58.4} & \textbf{33.7} & \textbf{30.9} \\ \hline
 \multirow{3}*{PointNet++} & 3D-Adv & 1.54 & 100 & 4.77 & 6.49 &  1.86 & 100 & 3.81 & 5.74 \\ & KNN & 2.77 & 100 & 5.35 &  7.91 &  2.63 & 100 & 5.45 &  7.29 \\ & AdvPC  & 4.81 &  99.2 & 28.2 & 18.9 & 4.81 &  99.2 & 28.3 & 18.9  \\   & AOF(ours)  &  \textbf{7.89} & 99.6 & \textbf{48.4} & \textbf{33.3} &  \textbf{8.62} & 99.9 & \textbf{48.4} & \textbf{34.3} \\
 \hline
 \multirow{4}*{PointConv} &  3D-Adv   & 1.45 &  6.58 & 100 &  3.02 &  1.32 & 6.92 & 100 &  3.33 \\ & KNN &  3.49 &  15.7 & 100 &  11.2 &  3.54 &  16.8 & 100 &  10.9 \\ & AdvPC  & 5.13 & 34.2 &  99.5 & 18.0 & 5.67 & 35.2 &  99.6 & 18.5 \\   & AOF(ours)  &  \textbf{6.85} & \textbf{50.0} & 99.9 & \textbf{25.5} &  \textbf{7.44} & \textbf{49.3} & 99.8 & \textbf{25.7} \\ \hline
 \multirow{4}*{DGCNN} &  3D-Adv  & 0.91 & 6.63 & 5.21 & 100 &  0.77 & 6.41 & 5.18 & 100 \\ & KNN  & 5.58 &  31.1 & 19.4 &  100 &  5.44 & 32.2 & 21.0 &  100 \\ & AdvPC & 7.44 & 60.0 & 44.5 &  93.7 & 8.08 & 60.4 & 44.3 &  93.6 \\   & AOF(ours)  &  \textbf{14.0} & \textbf{69.6} & \textbf{58.4} & 96.7 &  \textbf{16.6} & \textbf{69.2} & \textbf{58.7} & 96.7 \\
 \bottomrule
\end{tabular}
\label{tbl:transfer}
\vskip -0.1in
\end{table*}

% \textbf{The transferability of untargetted attack.} 
% We follow the experiment setup as~\cite{generatingadpoint,advpc}.
% by generating attacks
% using the constrained $l_\infty$ metric and measure their success rate at different
% norm-budgets $l_\infty$ taken to be in 0.18 and 0.45. 
% This range is chosen because it enables the attacks to reach $100\%$ success on the victim network, as well as offer an opportunity for transferability to other networks. 
We compare AOF against the SOTA baselines~\cite{generatingadpoint,knnattack, advpc} for $\epsilon_\infty=0.18$ and $\epsilon_\infty=0.45$. As discussed in Sec 4.1, we use ASR defined in Eq (\ref{eqn:asr}) as the real attack success rate. We report the transferability in Tabel~\ref{tbl:transfer}. From the results of Tabel~\ref{tbl:transfer}, it is clear that AOF consistently and significantly outperforms the baselines when transfer to other networks (up to nearly 70$\%$).

For reference, We also report the top-1 error rate following~\cite{advpc} in the appendix. As shown in Tabel~\ref{analysistransfer} and Tabel~\ref{tbl:transfer}, using the top-1 error rate to measure the transferability of the attacks is not fair. The key reason is that the clean accuracy of point cloud on the ModelNet40 test set cannot reach 100\%, therefore the point clouds that the victim model misclassified originally are also included when computing the top-1 error rate. As a result, the top-1 error rate cannot reflect the real gap between 3D adversarial attack algorithms. As shown in Tabel~\ref{tbl:transfer}, the gap between our AOF and AdvPC is more obvious than using the top-1 error rate in Table  ~\ref{analysistransfer}. While these experiments are for untargeted attacks, we perform similar experiments under targeted attacks and report the
results in supplement for reference and completeness.

% Moreover, we also visualized the adversarial samples crafted by AOF and AdvPC to analyze why AOF surpass AdvPC. As shown in Figure~\ref{visual_advpl}, our AOF tends to cause tiny deformation of original point clouds, rather than outliers and noise.

% \begin{table}[ht]
% \begin{center}
% \centering
% \caption{The classification accuracy (\%) of targeted attack $l_{\infty}$ is 0.18.}
% \label{target_acc}
% \footnotesize
% \begin{tabular}{c|c|c|c|c}
% % \hline
% \toprule
% Model  & PointNet & PointNet++ & DGCNN & PointConv \\
% \hline
% %   3D-Adv(ASR) & 0.16 & 0.73 & 99.76 & 0.69  \\
%   3D-Adv & 84.12 & 44.37 & 0.04 & 52.84  \\
% %  kNN(ASR)& 3.32 & 4.05 & 75.85 & 2.84  \\
%  kNN& 71.72 & 49.96 & 8.47 & 47.08  \\
% %  AdvPC(ASR)& 1.78 & 2.67 & 43.27 & 2.67  \\
%  AdvPC& 54.58 & 14.02 & 9.76 & 14.34  \\
% %  AOF(ASR)& 1.58 & 4.70 & 40.80 & 3.53  \\
%  AOF& \bf46.88 & \bf11.87 & 8.83 & \bf12.56  \\
% \bottomrule
% \end{tabular}
% \end{center}
% \vspace{-1ex}
% \end{table}

\subsection{AOF under Defense}
In order to explore the robustness of our AOF under various 3D adversarial defense algorithms, in this section, we first compare the attack success rate of AOF with other 3D adversarial attacks under various defense algorithms, then we compare the transferability of our AOF with other 3D adversarial attacks under various defenses, the results of which are in the appendix.

\textbf{Attack Success Rate under Defense.} We perform the SOTA 3D adversarial defenses to the adversarial point clouds crafted by our AOF and other 3D adversarial attacks. Several SOTA defenses: the Simple Random Sampling (SRS), Statistical Outlier Removal (SOR), DUP-Net~\cite{dupnet} and the IF-Defense~\cite{wu2020ifdefense} are selected. The results are showed in Table~\ref{compare_asr_with_alldefense}. The robustness against various defenses of our AOF are significantly better than all other 3D adversarial attacks showed in Table~\ref{compare_asr_with_alldefense}. The experimental results of targeted attacks under various defenses are reported in the appendix.

\begin{figure*}[htp!]
\vskip 0.2in
\begin{center}
% \resizebox{0.9\linewidth}{!}{
\centerline{\includegraphics[width=\textwidth]{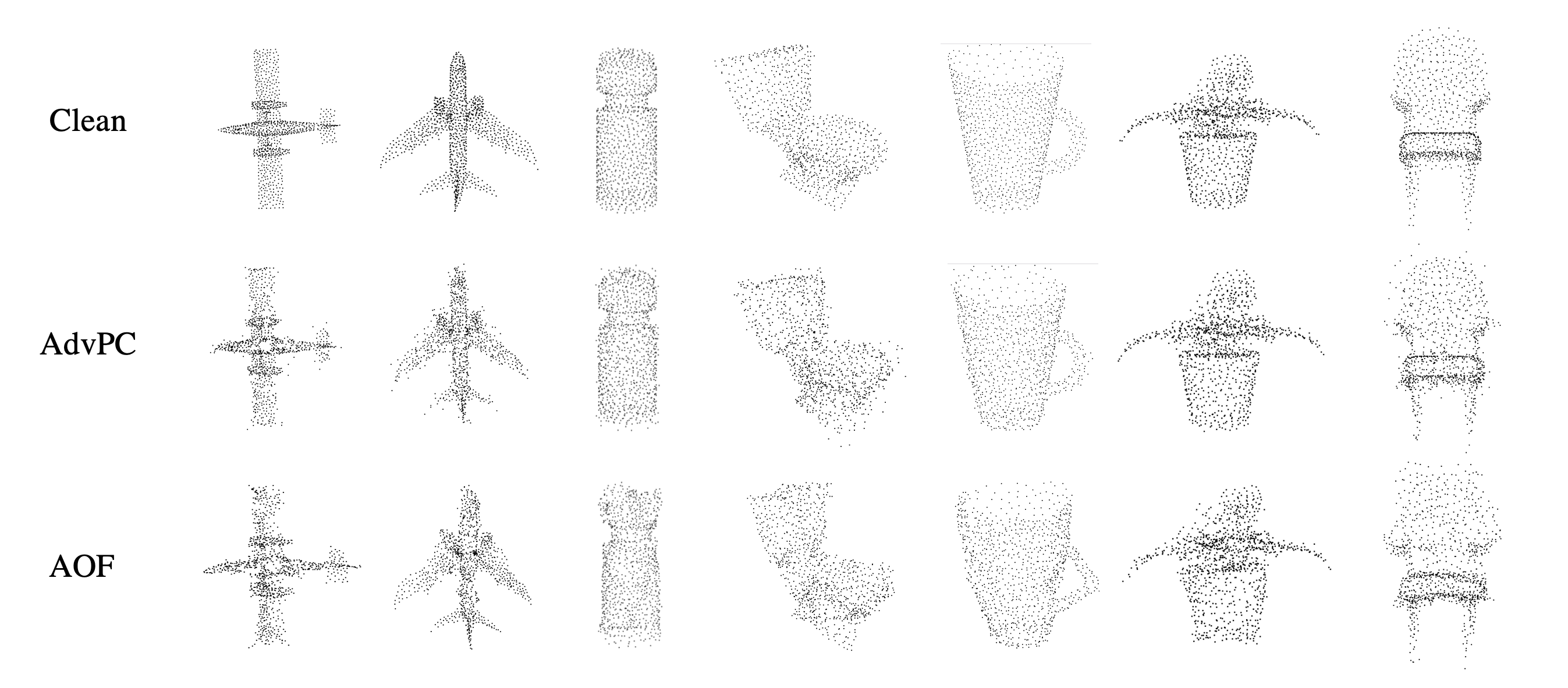}}
\caption{The visualization of adversarial samples}
\label{visual_advpl}
\end{center}
\vskip -0.2in
\end{figure*}

We visualized some adversarial samples crafted by AOF and AdvPC to analyze why AOF is more robust than AdvPC. As shown in Figure~\ref{visual_advpl}, our AOF tends to cause tiny deformation of original point clouds, rather than outliers. Deformation is usually harder to defend than outliers.

\begin{table*}[!t]
\caption{Attack success rate under various defense methods on PointNet++~\cite{pointnet++}, PointNet~\cite{pointnet}, DGCNN~\cite{dgcnn} and PointConv~\cite{wu2019pointconv}.The ConvNet-Opt, ONet-Remesh and ONet-Opt are three variants of IF-Defense.}
\label{compare_asr_with_alldefense}
\vskip 0.1in
\footnotesize
\centering
\setlength{\tabcolsep}{3.0mm}
\renewcommand\arraystretch{1.1}
% \caption{Attack success rate under various defense methods on PointNet++~\cite{pointnet++}, PointNet~\cite{pointnet}, DGCNN~\cite{dgcnn} and PointConv~\cite{wu2019pointconv}.The ConvNet-Opt, ONet-Remesh and ONet-Opt are three variants of IF-Defense.}
% \label{compare_asr_with_alldefense}
%\vspace{1mm}
%\resizebox{0.9\linewidth}{!}{
\begin{tabular}{c c c c c c c c c c}
\toprule
Victim models & Attacks & No defense &SRS& SOR & DUP-Net & ConvNet-Opt& ONet-Remesh& ONet-Opt\\
\midrule
\multirow{4}*{PointNet} &3D-Adv & 100 & 37.4 & 18.4 & 9.62& 5.58 & 10.8 & 7.4\\
&kNN  & 100 & 92.1 &  76.4&  30.5& 10.1&  12.9& 9.1\\
 &AdvPC & 93.7 & 89.6 & 53.6 &  23.1& 9.2& 12.6 & 7.6\\
 &AOF & 96.7 & \bf99.7 & \bf94.2 & \bf75.4& \bf27.9 &  \bf19.4 & \bf16.9\\

\midrule
\multirow{4}*{PointNet++} &3D-Adv & 100 & 65.9 & 27.3 & 22.9& 12.8 & 25.7 & 16.1\\
&kNN  & 100 & 79.6 &  \bf95.7&  80.6& 14.6&  25.5& 15.3\\
 &AdvPC & 93.7 & 86.8 & 79.0 &  72.1& 28.8& 34.5 & 24.7\\
 &AOF & 96.7 & \bf92.8 & 91.0 & \bf88.2& \bf45.6 &  \bf42.6 & \bf36.1\\
\midrule
\multirow{4}*{DGCNN} &3D-Adv & 100 & 29.5 & 23.7 & 23.2& 13.7 & 25.6 & 15.6\\
&kNN  & 100 & 59.5 &  24.5&  83.9& 18.7&  25.5& 17.4\\
 &AdvPC & 93.7 & 65.4 & 68.5 &  62.1& 26.3& 32.6 & 22.6\\
 &AOF & 96.7 & \bf75.8 & \bf79.3 & \bf76.0& \bf41.2 &  \bf39.5 & \bf32.5\\

\midrule
\multirow{4}*{PointConv} &3D-Adv & 100 & 42.4 & 48.0 & 37.3& 14.3 & 22.3 & 15.1\\
&kNN  & 100 & 73.9 &  \bf99.8&  95.9& 21.1&  24.2& 19.5\\
 &AdvPC & 93.7 & 80.5 & 94.0 &  88.1& 44.1& 30.9 & 37.5\\
 &AOF & 96.7 & \bf90.3 & 98.3 & \bf96.0& \bf59.6 &  \bf36.9 & \bf46.0\\
\bottomrule
\end{tabular}%}
\vskip -0.1in
\end{table*}

% \subsubsection{The attack transferability under defense}
% Motivated by the impressive performance of our AOF under defenses, we also test the transferability of untargeted attack under defense. The results are shown in  Table~\ref{compare_transferability_with_alldefense}. Our AOF is significantly better than SOTA 3D adversarial attacks under various defenses. The results are impressive because the previous attack algorithm is difficult to be transferred under the defense. The results further validate the effectiveness of our AOF.

\subsection{Discussion}
In this subsection, we explore the transferability of AOF when transfer to models trained with data augmentation. 
% Then we discuss an interesting phenomenon.
The complementing effect of AOF and spectral analysis of perturbation are discussed in the appendix due to page limits.
% \subsubsection{Complementing Effect of AOF}
% In principle, our AOF is compatible with other transfer-based 3D black-box adversarial attacks. Therefore, we can integrate the AOF with AdvPC. In this section, the victim model is PointNet, other experimental settings are the same with Sec 4.1. The experiment results are shown in Table~\ref{adPaof}. The AdvPC-AOF was constructed by applying AdvPC in AOF, which achieving the SOTA transferability. 

% \begin{table}[t]
% \caption{The attack sucess rate of AdvPC-AOF.}
% \label{adPaof}
% \vskip 0.15in
% \begin{center}
% \begin{small}
% % \begin{sc}
% \begin{tabular}{lcccr}
% \toprule
% Method & PointNet & PointNet++ & DGCNN & PointConv \\
% \midrule
% AdvPC    & 100 & 30.4 & 13.6 & 14.8\\
% AOF & 98.7 & 58.2 & 32.7 & 28.1\\
% AdvPC-AOF  & 100 & 67.1 & 44.6 & 35.9 \\
% \bottomrule
% \end{tabular}
% % \end{sc}
% \end{small}
% \end{center}
% \vskip -0.1in
% \end{table}

\begin{table}[hp!]
\caption{transferability drop rate (smaller numbers are better) of attacks when transfer to 3D networks trained with data augmentation. \textbf{Bold} numbers indicate the best.}
\label{trans_drop_aug}
\vskip 0.15in
\begin{center}
\begin{small}
%\begin{sc}
\setlength{\tabcolsep}{3.5mm}{
\begin{tabular}{lccccr}
\toprule
Method & PN & PN++ & PV & GCN \\
\midrule
% 3D-Adv    & 10.0(0.45) & 8.39(1.59) & 11.8(3.33) & 8.51(1.32)\\
% KNN & 12.2(3.01) & 11.5(4.45) & 17.0(8.70) & 19.2(13.0)\\
% AdvPC  & 13.6(4.68) & 23.7(18.4) & 39.0(34.0) & 24.8(19.5) \\
% AOF  & \textbf{17.5(9.13)} & \textbf{38.5(34.0)} & \textbf{52.3(48.8)} & \textbf{38.5(34.7)} \\

3D-Adv    & 50\% & 75\% & 36\% & 98\%\\
kNN & 46\% & 86\% & 55\% & 87\%\\
AdvPC  & 37\% & 69\% &  23\% & 79\% \\
AOF  & \textbf{34\%} & \textbf{51\%} & \textbf{16\%} & \textbf{64\%} \\
\bottomrule
\end{tabular}}
%\end{sc}
\end{small}
\end{center}
\vskip -0.1in
\end{table}

\textbf{Transferability with Data Augmentation.} The models trained with data augmentation are generally more robust than those without data augmentation~\cite{zhang2021pointcutmix,yun2019cutmix}. In this subsection, we explore the transferability of adversarial samples generated by AOF when transfer to models trained with data augmentation. The augmented models are trained on ModelNet40 with random scaling, random rotation, random shifting, random dropping and random point cloud jittering, the adversarial samples are generated by attacking DGCNN model trained without any data augmentation. Here, we pick $\epsilon_\infty = 0.18$, other experiment setting are the same as Sec 4.3. We denote the attck success rate on models without data augmentation as $\text{ASR}_{noaug}$ and the attack success rate on models with data augmentation as $\text{ARS}_{aug}$. Then, the drop rate of transferability can be expressed as 
\begin{align*} 
\label{eqn:xdecomp}
    \text{DR}_{ASR} = (\text{ASR}_{noaug} - \text{ASR}_{aug})/\text{ASR}_{noaug}.
\end{align*}
From Table~\ref{trans_drop_aug}, we can observe that adversarial samples generated by our method have a relatively smaller drop rate of transferability than other SOTA methods when transferred to models trained with data augmentation, which further verifies the robustness of AOF.

\section{Conclusions}
In this paper, we propose a new 3D adversarial attack, named AOF attack, which focuses on the LFC of point cloud to improve the transferability. Specifically, during the generation of adversarial point cloud, we optimize the LFC of point cloud by combining the the adversarial loss from the whole point cloud and its LFC. Extensive experiments on ModelNet40 validate that our AOF improves the transferability significantly compared to various SOTA 3D adversarial attack algorithms and is more robust to SOTA 3D adversarial defense algorithms. In the future, we plan to explore the imperceptibility of adversarial attack by restricting the perturbation of point cloud to HFC.

% \begin{figure}[ht]
% \vskip 0.2in
% \begin{center}
% \centerline{\includegraphics[width=\columnwidth]{icml_numpapers}}
% \caption{Historical locations and number of accepted papers for International
% Machine Learning Conferences (ICML 1993 -- ICML 2008) and International
% Workshops on Machine Learning (ML 1988 -- ML 1992). At the time this figure was
% produced, the number of accepted papers for ICML 2008 was unknown and instead
% estimated.}
% \label{icml-historical}
% \end{center}
% \vskip -0.2in
% \end{figure}

% % Note use of \abovespace and \belowspace to get reasonable spacing
% % above and below tabular lines.

% \begin{table}[t]
% \caption{Classification accuracies for naive Bayes and flexible
% Bayes on various data sets.}
% \label{sample-table}
% \vskip 0.15in
% \begin{center}
% \begin{small}
% \begin{sc}
% \begin{tabular}{lcccr}
% \toprule
% Data set & Naive & Flexible & Better? \\
% \midrule
% Breast    & 95.9$\pm$ 0.2& 96.7$\pm$ 0.2& $\surd$ \\
% Cleveland & 83.3$\pm$ 0.6& 80.0$\pm$ 0.6& $\times$\\
% Glass2    & 61.9$\pm$ 1.4& 83.8$\pm$ 0.7& $\surd$ \\
% Credit    & 74.8$\pm$ 0.5& 78.3$\pm$ 0.6&         \\
% Horse     & 73.3$\pm$ 0.9& 69.7$\pm$ 1.0& $\times$\\
% Meta      & 67.1$\pm$ 0.6& 76.5$\pm$ 0.5& $\surd$ \\
% Pima      & 75.1$\pm$ 0.6& 73.9$\pm$ 0.5&         \\
% Vehicle   & 44.9$\pm$ 0.6& 61.5$\pm$ 0.4& $\surd$ \\
% \bottomrule
% \end{tabular}
% \end{sc}
% \end{small}
% \end{center}
% \vskip -0.1in
% \end{table}

\bibliography{example_paper}
\bibliographystyle{icml2022}

%%%%%%%%%%%%%%%%%%%%%%%%%%%%%%%%%%%%%%%%%%%%%%%%%%%%%%%%%%%%%%%%%%%%%%%%%%%%%%%
%%%%%%%%%%%%%%%%%%%%%%%%%%%%%%%%%%%%%%%%%%%%%%%%%%%%%%%%%%%%%%%%%%%%%%%%%%%%%%%
% APPENDIX
%%%%%%%%%%%%%%%%%%%%%%%%%%%%%%%%%%%%%%%%%%%%%%%%%%%%%%%%%%%%%%%%%%%%%%%%%%%%%%%
%%%%%%%%%%%%%%%%%%%%%%%%%%%%%%%%%%%%%%%%%%%%%%%%%%%%%%%%%%%%%%%%%%%%%%%%%%%%%%%
\newpage
\appendix
\onecolumn
\section{Transferability with the Top-1 Error Rate}

\begin{table*}[ht!]
\caption{\small \textbf{Top-1 error rate of Attacks.} We use norm-budgets (max $\ell_\infty$ norm allowed in the perturbation) of $\epsilon_\infty = 0.18$ and $\epsilon_\infty = 0.45$ . All the reported results are the untargeted Top-1 error rate (higher numbers are better attacks). \textbf{Bold} numbers indicate the most transferable attacks. Our attack consistently achieves better transferability than the other attacks for all networks, especially on PointNet ~\cite{pointnet}. For reference, the classification accuracies on unperturbed samples for networks PointNet, PointNet++, PointConv and DGCNN are  89.3\%, 90.7\%, 90.6\%, and 91.0\%, respectively.
}
\label{analysistransfer}
\vskip 0.15in
\footnotesize
\setlength{\tabcolsep}{1.7mm} %
\renewcommand{\arraystretch}{1.1} %
\centering
% \resizebox{0.9\textwidth}{!}{%
\begin{tabular}{cc|cccc|cccc} 
\toprule
 &  &\multicolumn{4}{c|}{$\epsilon_\infty = 0.18$} &\multicolumn{4}{c}{$\epsilon_\infty = 0.45$}\\
\specialcell{Victim Networks} & Attacks  & PointNet & PointNet++ & PointConv & DGCNN  &PointNet & PointNet++  & PointConv & DGCNN   \\
\midrule
 \multirow{4}*{PointNet} & 3D-Adv  &  100 &  12.8 & 10.3 &  12.2 & 100 &  12.6 &  9.78 &  12.1 \\
 & KNN  & 100 &  17.9 & 11.4 &  17.4 & 100 &  17.5 &  11.3 &  17.0 \\
 & AdvPC  &  100 & 35.4 & 20.2 & 20.8 &  100 & 35.6 & 20.1 & 20.8 \\
  & AOF(ours)  &  99.9 & \textbf{60.7} & \textbf{37.6} & \textbf{32.7} &  100 & \textbf{61.2} & \textbf{38.6} & \textbf{35.5} \\ \hline
 \multirow{3}*{PointNet++} & 3D-Adv & 11.7 & 100 & 12.3 & 13.9 &  12.1 & 100 & 11.8 & 13.5 \\ & KNN & 12.4 & 100 & 12.9 &  14.4 &  12.3 & 100 & 13.2 &  13.9 \\ & AdvPC  & 14.4 &  99.2 & 33.0 & 24.7 & 14.4 &  99.2 & 33.1 & 24.8  \\   & AOF(ours)  &  \textbf{16.9} & 99.6 & \textbf{51.6} & \textbf{37.7} &  \textbf{17.6} & 99.9 & \textbf{51.7} & \textbf{38.6} \\  
 \hline
 \multirow{4}*{PointConv} &  3D-Adv   & 11.8 &  14.3 & 100 &  10.9 &  11.8 & 14.7 & 100 &  11.1 \\ & KNN &  13.3 &  22.7 & 100 &  17.3 &  13.2 &  23.4 & 100 &  17.0 \\ & AdvPC  & 15.1 & 39.1 &  99.5 & 23.7 & 15.5 & 39.9 &  99.6 & 23.9 \\   & AOF(ours)  &  \textbf{16.5} & \textbf{53.4} & 99.9 & \textbf{30.6} &  \textbf{17.0} & \textbf{53.0} & 99.8 & \textbf{30.8} \\ \hline
 \multirow{4}*{DGCNN} &  3D-Adv  & 11.1 & 13.6 & 12.4 & 100 &  10.9 & 13.4 & 12.7 & 100 \\ & KNN  & 15.0 &  35.4 & 25.6 &  100 &  14.8 & 37.3 & 27.2 &  100 \\ & AdvPC & 16.6 & 62.2 & 47.8 &  93.7 & 16.8 & 62.4 & 47.5 &  93.6 \\   & AOF(ours)  &  \textbf{22.3} & \textbf{70.6} & \textbf{60.8} & 96.7 &  \textbf{24.6} & \textbf{70.1} & \textbf{61.1} & 96.7 \\
 \bottomrule
\end{tabular}
\vskip -0.1in
\end{table*}

\section{The Transferability of Targeted Attacks}
As shown in Table~\ref{target_trans_asr}, the transferability of targeted attack are much lower than untargeted attack, and our AOF has a comparable targeted transferability with kNN. Improving the transferability of targeted attack is a challenging but promising task. From Table~\ref{target_acc}, in terms of misclassification rate, our AOF still outperforms the other baselines when transfer to different networks.
% Here, we show the transferanility of targeted attacks using to DGCNN with $\epsilon_\infty=0.18$. We represent the transferability and the top-1 error rate under targeted attacks in Table~\ref{target_trans_asr} and Table~\ref{target_acc} respectively.
% The targeted attacks pose a greater threat than untargeted attacks. For example, if the 3D point cloud classification model that was implement in a self-driving car was attacked by the targeted attack, if the attacker aims to misclassify the people class to the chair class, this will cause serious traffic accidents. 

% We therefore firstly investigated the transferability of targeted attack in 3D point cloud. The results are shown in Table~\ref{target_acc} and Table~\ref{target_trans_asr}, the victim model is DGCNN, the transferability of targeted attack are much lower than untargeted attack, indicating improving the transferability of targeted attack is a challenging but promising task. 

\begin{table}[h!]
\caption{\textbf{Transferability of Targeted Attacks}. Here, we pick $\epsilon_{\infty}=0.18$, $m=100$, $\gamma=0.25$ and select DGCNN as victim model. Other settings are the same as Sec 4.1. We use the targeted attack success rate (\textbf{higher} indicates better attacks) to evaluate the transferability.}
\label{target_trans_asr}
\vskip 0.15in
\begin{center}
%\setlength{\tabcolsep}{5.0mm}
%\begin{small}
%\begin{sc}
\begin{tabular}{lccccr}
\toprule
Model  & PointNet & PointNet++ & DGCNN & PointConv \\
\midrule
3D-Adv & 0.16 & 0.73 & \bf99.76 & 0.69  \\
%   3D-Adv(ACC) & 84.12 & 44.37 & 0.04 & 52.84  \\
kNN& \bf3.32 & \bf4.05 & 75.85 & 3.16  \\
%  kNN(ACC)& 71.72 & 49.07 & 8.06 & 47.08  \\
AdvPC& 1.01 & 1.90 & 76.56 & 1.70  \\
%  AdvPC(ACC)& 60.04 & 20.81 & 3.77 & 20.00  \\
AOF& 2.39 & 3.28 & 77.17 & \bf3.52  \\
%  AOF(ACC)& 53.60 & 13.56 & 3.85 & 13.16  \\
\bottomrule
\end{tabular}
%\end{sc}
%\end{small}
\end{center}
\vskip -0.1in
\end{table}

\begin{table}[h!]
\caption{\textbf{Classification Accuracy} (smaller indicates better results) of different models on targeted adversarial samples. Here, we pick $\epsilon_{\infty}=0.18$, $m=100$, $\gamma=0.25$ and select DGCNN as victim model. Other settings are the same as Sec 4.1. In terms of misclassification rate, our AOF outperforms the other baselines when transferred to different networks.}
\label{target_acc}
\vskip 0.15in
\begin{center}
%\setlength{\tabcolsep}{5.0mm}
%\begin{small}
%\begin{sc}
\begin{tabular}{lccccr}
\toprule
Model  & PointNet & PointNet++ & DGCNN & PointConv \\
\midrule
3D-Adv & 84.12 & 44.37 & \bf0.04 & 52.84  \\
kNN & 71.72 & 49.07 & 8.06 & 47.08  \\
AdvPC & 60.04 & 20.81 & 3.77 & 20.00  \\
AOF & \bf53.60 & \bf13.56 & 3.85 & \bf13.16  \\
\bottomrule
\end{tabular}
%\end{sc}
%\end{small}
\end{center}
\vskip -0.1in
\end{table}

\section{The Attack Transferability under Defense.}
Motivated by the impressive performance of our AOF under defenses, we also test the transferability of untargeted attack under defense. The results are shown in  Table~\ref{compare_transferability_with_alldefense}. Our AOF is significantly better than SOTA 3D adversarial attacks under various defenses. The results are impressive because the previous attack algorithm is difficult to be transferred under the defense. The results further validate the effectiveness of our AOF.

\begin{table*}[!t]
\caption{Averaged transferability under various defense methods on Pointnet++~\cite{pointnet++}, Pointnet~\cite{pointnet}, DGCNN~\cite{dgcnn} and PointConv~\cite{wu2019pointconv}. The ConvNet-Opt, ONet-Remesh and ONet-Opt are three variants of IF-Defense.}
\label{compare_transferability_with_alldefense}
\vskip 0.15in
\footnotesize
\centering
\setlength{\tabcolsep}{3.0mm}
\renewcommand\arraystretch{1.1}
%\resizebox{0.7\linewidth}{!}{
\begin{tabular}{c c c c c c c c c c}
\toprule
Victim models & Attacks & No defense &SRS& SOR & DUP-Net & ConvNet-Opt& ONet-Remesh& ONet-Opt\\
\midrule
\multirow{4}*{PointNet} &3D-Adv & 3.8 & 23.2 & 4.40 & 7.40 & 10.6 & 23.3 & 13.1\\
&kNN  & 8.5 & 27.4 &  6.2&  9.4& 11.1&  23.7& 13.7\\
 &AdvPC & 19.6 & 42.3 & 11.1 &  14.3& 12.5& 26.0 & 15.0\\
 &AOF & \bf39.6 & \bf63.9 & \bf33.5 & \bf35.5& \bf25.0 &  \bf35.9 & \bf25.3\\

\midrule
\multirow{4}*{PointNet++} &3D-Adv & 4.3 & 16.0 & 5.60 & 9.40 & 8.80 & 19.4 & 10.5\\
&kNN  & 5.3 & 15.4 &  5.8&  9.1& 8.3&  19.2& 10.7\\
 &AdvPC & 17.3 & 29.0 & 16.0 &  19.8& 16.1& 23.5 & 17.1\\
 &AOF & \bf29.8 & \bf40.5 & \bf28.2 & \bf30.4& \bf23.9 &  \bf28.7 & \bf22.5\\
\midrule
\multirow{4}*{PointConv} &3D-Adv & 3.7 & 22.3 & 6.00 & 8.80 & 9.60 & 20.2 & 12.2\\
&kNN  & 10.1 & 26.9 &  13.7&  16.2& 10.0&  21.2& 12.7\\
 &AdvPC & 19.1 & 39.3 & 23.7 &  24.8& 16.7& 25.9 & 17.1\\
 &AOF & \bf27.4 & \bf46.0 & \bf33.0 & \bf34.3& \bf22.5 &  \bf29.3 & \bf21.3\\

\midrule
\multirow{4}*{DGCNN} &3D-Adv & 4.3 & 25.2 & 8.70 & 11.3& 10.1 & 19.4 & 11.9\\
&kNN  & 18.7 & 34.5 &  23.1&  26.4& 9.9&  19.8& 12.4\\
 &AdvPC & 37.3 & 48.9 & 35.3 &  36.1& 24.4& 27.0 & 23.0\\
 &AOF & \bf47.3 & \bf55.9 & \bf45.2 & \bf46.6& \bf35.0 &  \bf39.9 & \bf30.9\\
\bottomrule
\end{tabular}
\vskip -0.1in
\end{table*}

\section{The Performance of Targeted Attacks under Defense.}
To evaluate the performance of targeted attacks under various defenses. Firstly, We adopt 1 - accuracy to evaluate the untargeted robustness of attacks. From Table ~\ref{tab:targeted_defense_acc}, it is clear that AOF outperforms other SOTA attacks by a large margin. Then we assess the targeted robustness of attacks using targeted attack success rate. As shown in Table ~\ref{tab:targeted_defense_asr}, our AOF outperforms other SOTA attacks under IF-Defense and reaches the second best position under SRS, SOR and DUP defenses.

\begin{table}[h!]
\caption{In this table, we evaluate the performance of targeted attacks under different defenses using \textbf{1 - accuracy} (higher indicates better attacks)). We pick $\epsilon_{\infty}=0.18$, $m=100$, $\gamma=0.25$ and select DGCNN as victim model. Other settings are the same as Sec 4.1. Our AOF consistently and significantly outperforms other SOTA attacks. The ConvNet-Opt, ONet-Remesh and ONet-Opt are three variants of IF-Defense.}
\label{tab:targeted_defense_acc}
\vskip 0.15in
\begin{center}
\setlength{\tabcolsep}{4.0mm}
%\begin{small}
%\begin{sc}
\begin{tabular}{lccccr}
\toprule
Defenses  & 3D-Adv & kNN & AdvPC & AOF \\
\midrule
No defense & \bf99.76 & 91.94 & 96.23 & 96.15  \\
SRS & 54.42 & 71.11 & 88.02 & \bf91.26  \\
SOR & 27.92 & 54.78 & 75.47 & \bf85.63  \\
DUP-Net & 29.58 & 51.66 &71.90 & \bf83.16  \\
ConvNet-Opt & 21.76 & 22.85 & 45.59 & \bf61.38  \\
ONet-Remesh & 30.27 & 30.19 & 49.51 & \bf58.46  \\
ONet-Opt & 24.51 & 25.24 & 41.34 & \bf51.34  \\
\bottomrule
\end{tabular}
%\end{sc}
%\end{small}
\end{center}
\vskip -0.1in
\end{table}

\begin{table}[h!]
\caption{The attack success rate (\textbf{higher} indicates better attacks) of targeted attacks under different defenses. We pick $\epsilon_{\infty}=0.18$, $m=100$, $\gamma=0.25$ and select DGCNN as victim model. Other settings are the same as Sec 4.1. Our AOF outperforms other SOTA attacks under IF-Defense and reaches the second best position under SRS, SOR and DUP defenses. The ConvNet-Opt, ONet-Remesh and ONet-Opt are three variants of IF-Defense.}
\label{tab:targeted_defense_asr}
\vskip 0.15in
\begin{center}
\setlength{\tabcolsep}{4.0mm}
%\begin{small}
%\begin{sc}
\begin{tabular}{lccccr}
\toprule
Defenses  & 3D-Adv & kNN & AdvPC & AOF \\
\midrule
No defense & \bf99.76 & 75.85 & 76.56 & 77.17  \\
SRS & 3.04 & \bf13.98 & 6.88 & 7.09  \\
SOR & 1.54 & \bf22.00 & 8.54 & 14.05  \\
DUP-Net & 1.38 & \bf13.25 &7.04 & 11.38  \\
ConvNet-Opt & 0.57 & 2.84 & 1.09 & \bf4.66  \\
ONet-Remesh & 0.85 & 1.62 & 1.05 & \bf1.98  \\
ONet-Opt & 0.65 & 1.94 & 1.01 & \bf2.91  \\
\bottomrule
\end{tabular}
%\end{sc}
%\end{small}
\end{center}
\vskip -0.1in
\end{table}

\section{Complementing Effect of AOF}
In principle, our AOF is compatible with other transfer-based 3D black-box adversarial attacks. We can integrate the AOF with AdvPC. In this section, the victim model is PointNet, and we pick $\epsilon_\infty=0.18$. Other experimental settings are the same as Sec 4.1. The experiment results are shown in Table~\ref{adPaof}. The AdvPC-AOF constructed by applying AdvPC to AOF, achieves the SOTA transferability. 

\begin{table*}[h!]
\caption{The attack sucess rate of AdvPC-AOF.}
\label{adPaof}
\vskip 0.15in
\begin{center}
%\setlength{\tabcolsep}{5.0mm}
%\begin{small}
%\begin{sc}
\begin{tabular}{lccccr}
\toprule
Method & PointNet & PointNet++ & DGCNN & PointConv \\
\midrule
AdvPC    & 100 & 30.4 & 13.6 & 14.8\\
AOF & 98.7 & 58.2 & 32.7 & 28.1\\
AdvPC-AOF  & 100 & \bf67.1 & \bf44.6 & \bf35.9 \\
\bottomrule
\end{tabular}
%\end{sc}
%\end{small}
\end{center}
\vskip -0.1in
\end{table*}

\section{Spectral Analysis of Perturbation}
\begin{figure}[ht!]
\vskip 0.2in
\begin{center}
\centerline{\includegraphics[width=0.75\columnwidth]{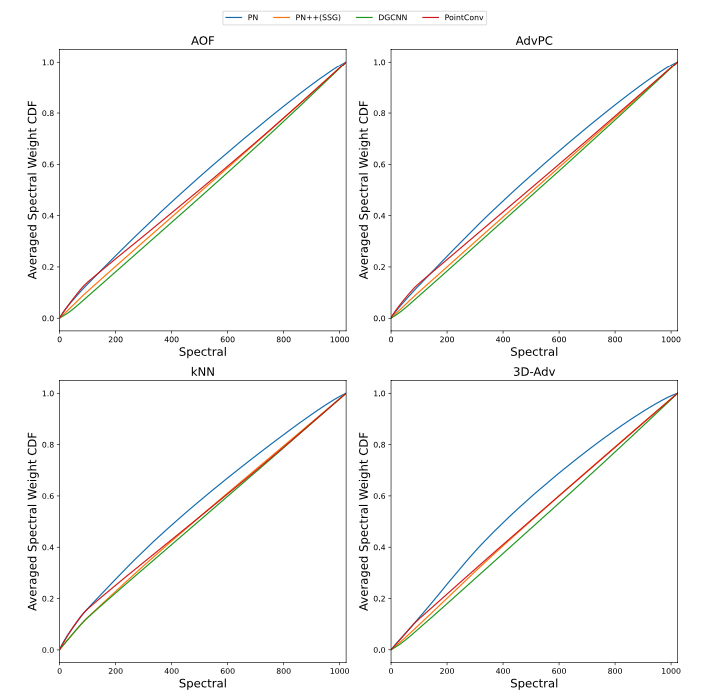}}
\caption{
% \text{Sensitivity Analysis}: Studying t
The spectral analysis of adversarial perturbation.}
\label{spanalysis}
\end{center}
\vskip -0.2in
\end{figure}

We found that the transferability to PointNet of attack algorithm is relatively lower than other networks. In order to briefly explore the cause of this phenomenon, we do a simple spectral analysis of the adversarial perturbation from different networks. We first get the spectral weight of perturbation by projecting the perturbation to the eigenvectors of Laplacian matrix of original point cloud. Then we calculate the cumulative distribution of spectral weights for each adversarial sample. Finally, we compute the average of the cumulative distributions of all adversarial samples to get the averaged spectral weight cdf for a specific victim model in Figure~\ref{spanalysis}. As shown in Figure~\ref{spanalysis}, The spectral weight cdf curve of PointNet is obviously different from that of PointNet++, PointConv and DGCNN. The perturbation of PointNet has a higher proportion of LFC than PointNet++ and DGCNN, maybe that's the reason why it is harder to transfer to PointNet than PointNet++, PointConv and DGCNN.

%%%%%%%%%%%%%%%%%%%%%%%%%%%%%%%%%%%%%%%%%%%%%%%%%%%%%%%%%%%%%%%%%%%%%%%%%%%%%%%
%%%%%%%%%%%%%%%%%%%%%%%%%%%%%%%%%%%%%%%%%%%%%%%%%%%%%%%%%%%%%%%%%%%%%%%%%%%%%%%

\end{document}